\documentclass[sigconf]{acmart}
\AtBeginDocument{%
  }

\usepackage{multirow} 
\usepackage{booktabs} 
\usepackage{makecell}  
\usepackage{graphicx}
\usepackage{subcaption}
\usepackage{graphicx}
\usepackage{algorithm}
\usepackage{algpseudocode}

\setcopyright{rightsretained}
\copyrightyear{2026}
\acmYear{2026}
\acmConference[WWW '26]{The Web Conference 2026}{April 13--17, 2026}{Dubai, United Arab Emirates}

\begin{document}

\title{Fine-Grained Traceability for Transparent ML Pipelines}

\author{Liping Chen}
\affiliation{%
  % \department{School of Computing Technologies}
  \institution{RMIT University}
  \city{Melbourne}
  \state{VIC}
  \country{Australia}
}
\email{lp.chen@ieee.org}

\author{Mujie Liu}
\affiliation{%
  % \department{Institute of Innovation, Science and Sustainability}
  \institution{Federation University Australia} % Australia
  \city{Ballarat}
  \state{VIC}
  \country{Australia}
}
\email{mujie.liu@ieee.org}

\author{Haytham Fayek}
\authornote{Corresponding author.}
\affiliation{%
  % \department{School of Computing Technologies}
  \institution{RMIT University}
  \city{Melbourne}
  \state{VIC}
  \country{Australia}
}
\email{haytham.fayek@ieee.org}

\renewcommand{\shortauthors}{Chen et al.}

\begin{abstract} 
Modern machine learning systems are increasingly realised as multi-stage pipelines, yet existing transparency mechanisms typically operate at a model level: they describe what a system is and why it behaves as it does, but not how individual data samples are operationally recorded, tracked, and verified as they traverse the pipeline. This absence of verifiable, sample-level traceability leaves practitioners and users unable to determine whether a specific sample was used, when it was processed, or whether the corresponding records remain intact over time. We introduce FG-Trac, a model-agnostic framework that establishes verifiable, fine-grained sample-level traceability throughout machine learning pipelines. FG-Trac defines an explicit mechanism for capturing and verifying sample lifecycle events across preprocessing and training, computes contribution scores explicitly grounded in training checkpoints, and anchors these traces to tamper-evident cryptographic commitments. The framework integrates without modifying model architectures or training objectives, reconstructing complete and auditable data-usage histories with practical computational overhead. Experiments on a canonical convolutional neural network and a multimodal graph learning pipeline demonstrate that FG-Trac preserves predictive performance while enabling machine learning systems to furnish verifiable evidence of how individual samples were used and propagated during model execution.
\end{abstract}
%%




%\received{30 November 2025}

\maketitle

\section{Introduction}

Modern machine learning systems are no longer defined solely by model architectures, but by complex, multi-stage pipelines whose behaviour emerges from data ingestion, preprocessing, batching, gradient computation, checkpointing, and deployment~\cite{schlegel2025capturing}. 
These pipelines govern how data flows, how representations are learned, and how predictions are produced, yet their complexity obscures the involvement of individual samples in these processes. Crucially, current machine learning pipelines lack operational transparency at the sample level: there is no principled mechanism to determine what happens to each data point as it traverses these stages. Although every training sample participates directly in these operations and shapes the final model, existing pipelines offer virtually no visibility into how individual samples are consumed, transformed, or propagated~\cite{mora2021traceability,spoczynski2025atlas}. In practice, there is no reliable record of when a sample was used, which updates it influenced, or whether it contributed to a specific prediction~\cite{grafberger2023provenance}. 

This absence of verifiable, sample-level traceability is not a usability issue, but a structural limitation. It undermines accountability, hinders post-hoc forensic reconstruction, and prevents practitioners from verifying lawful data usage, particularly in regulated domains such as healthcare, finance, and security~\cite{mu2023data,huang2025instance}. While not all machine learning applications require such guarantees, pipelines in high-risk domains cannot be deployed responsibly without them. This motivates a systematic solution for sample-level traceability that is independent of any specific model architecture.

Prior work on influence estimation offers partial proxies for understanding how individual samples affect model outputs. Methods such as Influence Functions~\cite{koh2017understanding}, TracIn and TracInCP~\cite{pruthi2020estimating}, and Data Shapley~\cite{jia2019towards} estimate the marginal contribution of training examples to predictions. However, these approaches do not provide operational or verifiable sample-level traceability grounded in the actual execution of the machine learning pipeline, for three reasons:
(i) They operate entirely outside the training pipeline, producing estimates that cannot be validated against the actual optimisation trajectory of the model or the proper sequence of sample usage.
(ii) They return isolated scalar scores that provide little temporal or semantic context, offering no insight into when, where, or through which checkpoints a sample supported or opposed a prediction~\cite{koh2017understanding}.
(iii) They lack integrity guarantees: their outputs can be regenerated, modified, or selectively omitted without evidence, and no verifiable record ties influence values to concrete training events~\cite{schlegel2025capturing}.
Consequently, existing estimators function as retrospective explanations rather than verifiable operational records and cannot satisfy accountability requirements in regulated or high-stakes pipelines. They quantify influence, but do not constitute a mechanism for traceability and cannot operationalise accountability in machine learning systems.

To address this, we introduce FG-Trac, a fine-grained, model-agnostic framework that provides verifiable, end-to-end sample-level traceability throughout the machine learning pipeline. FG-Trac is designed to address the above limitations through three aligned capabilities.
(i) It instruments the pipeline to record sample-level lifecycle events, including preprocessing, batching, gradient computation, and evaluation, ensuring that influence assessments are grounded in the model’s actual execution rather than retrospective approximations.
(ii) It computes structured, checkpoint-resolved influence scores via a pipeline-integrated variant of TracInCP, producing temporally indexed traces that capture how each sample contributes, positively or negatively, across the course of training.
(iii) It aggregates all sample-level events and influence traces into Merkle-root digests anchored to an immutable ledger, providing tamper-evident integrity and auditability that prior estimators fundamentally lack. 
Notably, FG-Trac is a plug-and-play module that can be integrated into diverse machine learning pipelines with minimal modification. It provides verifiable, sample-level traceability throughout machine learning pipelines, across model classes, datasets, and application domains, thereby complementing rather than replacing existing influence estimators. Experiments across heterogeneous training settings demonstrate its broad applicability, while evaluations on two real-world medical datasets confirm its ability to provide reliable, fine-grained sample-level provenance. On these medical datasets, FG-Trac not only preserves baseline predictive performance but also enables machine learning systems to answer operational questions that were previously inaccessible: \emph{Which samples were used and when? How did they affect the model’s behaviour? Can their influence be independently verified?} Our main contributions can be summarised as follows:
\begin{itemize}
    \item We formalise the problem of operational, verifiable sample-level traceability and introduce FG-Trac, a general framework that couples provenance tracking directly with machine learning pipeline execution, clarifying what must be recorded and validated to capture how samples are used and how they influence model updates.
    \item We design a verifiable, tamper-evident influence estimation mechanism by integrating a structured, checkpoint-resolved variant of TracInCP and securing all sample-level events with Merkle-root digests anchored to an immutable ledger.
    \item We demonstrate FG-Trac’s generality and practicality through representative evaluations on heterogeneous machine learning pipelines and two real-world medical datasets, showing preserved predictive performance and reliable sample-level provenance.
    
\end{itemize}

% Main result
% so what
% We instantiate FG-Trac across diverse pipelines and datasets. Experiments demonstrate that the framework reconstructs complete sample-level usage histories, produces stable and interpretable influence estimates, and preserves predictive performance with practical overhead. Together, these results show that fine-grained, verifiable traceability can be realised as a general, pipeline-level capability, complementing existing documentation artefacts with operational, sample-level accountability for auditing and responsible ML deployment.

\section{Related Work}

\subsection{Transparency Artefacts for Machine Learning Systems}

Recent work has advanced transparency in machine learning by documenting how models, datasets, and pipelines should be reported and contextualised. Model Cards~\cite{mitchell2019model} formalise disclosure around model usage, evaluation protocols, and demographic considerations. Data Cards~\cite{pushkarna2022data} complement this by documenting dataset provenance, curation decisions, and transformation workflows. Use Case Cards~\cite{hupont2024use} broaden the scope to deployment contexts and operational risks, reflecting the increasing importance of governance beyond model performance. DAG Cards~\cite{tagliabue2021dag} extend this paradigm to pipeline-level structure and behavioural testing, marking a shift from model-centric to pipeline-centric transparency practices. These artefacts expand what can be made visible in machine learning systems, yet they primarily describe system configurations rather than the operational histories of individual samples as they flow through pipelines.

\subsection{Traceability in Machine Learning Systems}

Traceability in machine learning is conceptually distinct from explainability. Whereas explainability concerns whether and how a model’s decisions can be interpreted, traceability addresses whether the operational use of data within a machine learning pipeline can be reconstructed~\cite{chander2025toward}. Most existing work on traceability is grounded in data provenance, documenting how datasets, models, and artefacts are created, transformed, and reused over time. Systems such as ModelDB~\cite{vartak2016modeldb}, PROV-AGENT~\cite{souza2025prov}, OpenML~\cite{casalicchio2019openml}, and MLflow~\cite{chen2020developments} exemplify this line of research by recording metadata, configurations, and artefact versions to support reproducibility and pipeline-level auditing. Beyond lifecycle provenance, related efforts examine behavioural visibility in specific domains, including interaction-level logging, event-dependency analysis for security, and semantic tracing in NLP workflows~\cite{10.1145/3583780.3615998,liu2025trace2vec,yanez2025text}. Taken together, these approaches improve visibility into pipeline structure and inference-time behaviour but provide limited insight into how individual samples participate in optimisation or shape training dynamics.

A related line of work focuses on ensuring the integrity of execution records through verifiable logging mechanisms. Techniques such as blockchain-based logging, Merkle-tree commitments, and ledger anchoring~\cite{li2023graph,10185966} guarantee that once records are produced, they cannot be silently modified or selectively removed, strengthening trust in recorded histories and supporting audit and compliance requirements. However, these mechanisms primarily address how records are protected, rather than which operational events must be captured to enable meaningful traceability. As a result, existing systems either provide detailed provenance without verifiability guarantees or enforce immutability over logs that omit sample-level operational histories, leaving the traceability of model training incomplete at the level of individual data samples.

\subsection{Influence and Contribution Estimation}
A body of work studies how individual samples affect model predictions. Methods such as Influence Functions~\cite{koh2017understanding}, TracIn and TracInCP~\cite{pruthi2020estimating}, Data Shapley~\cite{jia2019towards}, and Representer Point Selection~\cite{sui2021representer} quantify the importance of training examples by estimating their marginal contribution to learned parameters or outputs. These approaches offer insight into which samples support or interfere with specific predictions, and practical variants such as TracInCP make influence estimation tractable in modern deep learning settings. However, influence estimation is not designed as an operational record: values are computed retrospectively and remain decoupled from when, where, and under which model state samples are used during training, nor do they support verification against concrete training events.

\section{The FG-Trac Framework}

\begin{figure*}[t]
\centering
\includegraphics[width=\textwidth]{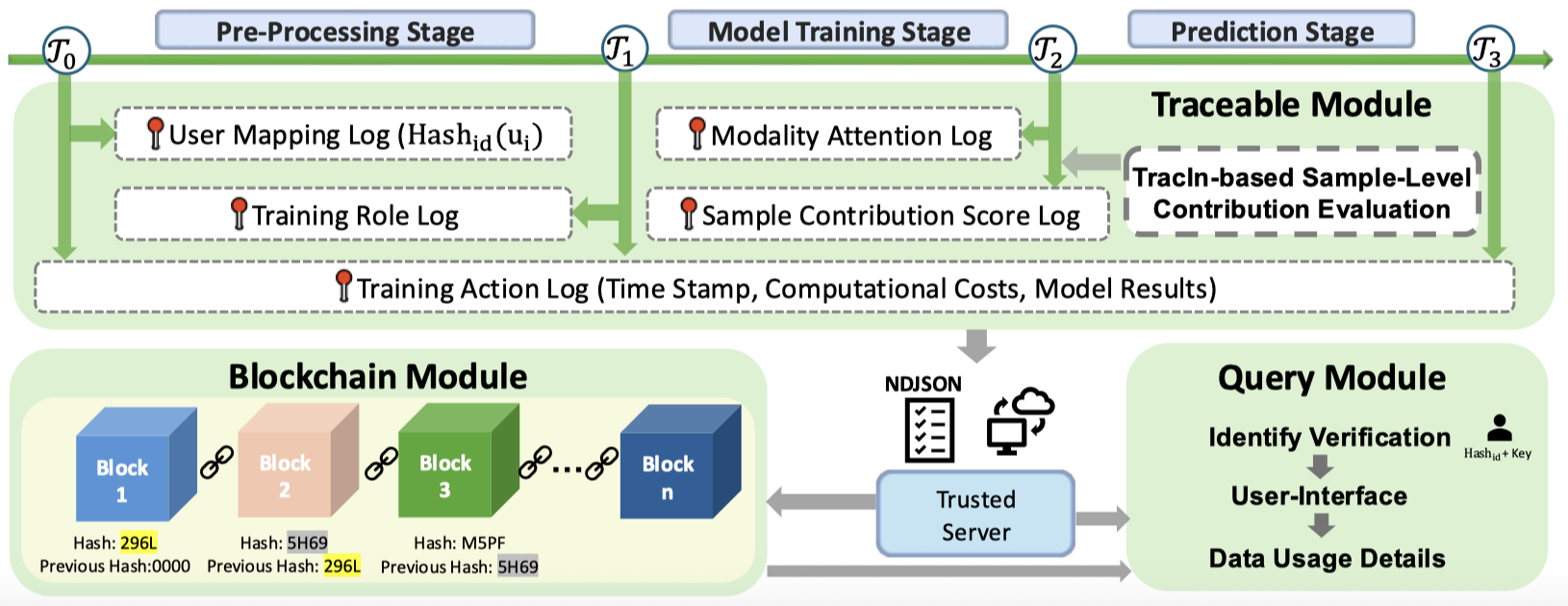}
\caption{Overview of the FG-Trac framework.}
\Description{An overview of the FG-Trac framework showing the Traceable Module, Blockchain Module, and Query Module aligned with preprocessing, training, and prediction stages of a machine learning pipeline.}
\label{fig:framework}
\end{figure*}

FG-Trac operationalises verifiable, sample-level traceability through system-level instrumentation and auditable logging mechanisms that integrate directly into a machine learning pipeline. The framework comprises three interconnected components: the \textit{Traceable Module}, the \textit{Blockchain Module}, and the \textit{Query Module}. Each component addresses a core requirement of accountable, end-to-end sample tracking. Figure~\ref{fig:framework} illustrates how FG-Trac aligns these modules with key pipeline stages, from preprocessing ($\mathcal{T}_0$–$\mathcal{T}_1$) to model training ($\mathcal{T}_1$–$\mathcal{T}_2$) and prediction ($\mathcal{T}_2$–$\mathcal{T}_3$).

\subsection{Traceable Module}

To address the first question: \emph{Which samples were used and when?} We introduce the Traceable Module, which instruments the pipeline to produce fine-grained, sample-centric operational logs across all stages (as shown in Figure~\ref{fig:framework}). Rather than logging only coarse pipeline stages, FG-Trac records five types of operational events:

\begin{itemize}
    \item \textbf{User Mapping Log ($\mathcal{T}_0$).} At ingestion time, each user or subject $u_i$ is mapped to a pseudonymous identifier
    \begin{equation}
        h_i = \mathrm{SHA256}(u_i),
    \end{equation}
    implemented as Hash\_id($u_i$) in our prototype. This log links internal records to hashed identifiers without storing raw identities.

    \item \textbf{Training Role Log ($\mathcal{T}_0\!\rightarrow\!\mathcal{T}_1$).} When the dataset is split, FG-Trac records the role of each sample (e.g., training/ validation/ test) together with its pseudonymous identifier. This enables later queries about \emph{how} a given sample participated in model development.

    \item \textbf{Modality Attention Log ($\mathcal{T}_1\!\rightarrow\!\mathcal{T}_2$).} During model training, FG-Trac optionally records summarised modality-level attention patterns (e.g., for imaging vs.\ clinical text in a multimodal graph neural network (GNN)) for each sample or batch. These logs provide interpretable signals about which modalities were emphasised when the model saw a given sample, without storing raw activations.

    \item \textbf{Sample Contribution Score Log ($\mathcal{T}_1\!\rightarrow\!\mathcal{T}_2$):} At selected optimisation checkpoints, FG-Trac records signed contribution scores produced by its TracInCP-based~\cite{pruthi2020estimating} estimator (see below) for relevant pairs of target samples (i.e., queried examples whose predictions are being explained) and training samples. These records summarise whether each training sample supports or interferes with target predictions, and form the core of FG-Trac’s influence-aware traceability.

    \item \textbf{Training Action Log ($\mathcal{T}_0$--$\mathcal{T}_3$).} Finally, FG-Trac maintains a global action log that records timestamps, computational costs, model versions, and key training events. This log ties the sample-level records to concrete runs and model states.
\end{itemize}

Together, these logs provide a fine-grained, sample-centric footprint across preprocessing, training and fine-tuning, and prediction, which underpins the verifiable traces exposed by the Query Module.

\medskip
\noindent\textbf{Contribution Estimation.}
\label{sec:contribution}
To answer the second introductory question: \emph{How did these samples affect the model’s behaviour?} FG-Trac integrates a checkpoint-based contribution estimator, TracInCP~\cite{pruthi2020estimating}, into the training process. Rather than computing influence along the full optimisation trajectory, FG-Trac evaluates gradient alignment only at a selected set of discrete checkpoints $\mathcal{C}$, reducing overhead while preserving interpretability. Let $\theta_c$ denote the model parameters saved at checkpoint $c$, and let $\ell(\theta, s)$ be the loss incurred by sample $s$. For a target sample $s_i$ and a candidate sample $s_j$, the checkpointed influence score is computed as
\begin{equation}
    I_{\mathrm{CP}}(s_j \rightarrow s_i)
    =
    \sum_{c\in\mathcal{C}}
    \nabla_{\theta}\ell(\theta_c, s_i)^{\!\top}
    \nabla_{\theta}\ell(\theta_c, s_j),
\end{equation}
where the gradient inner product measures whether $s_j$ supports or interferes with $s_i$ across optimisation states. Positive values indicate that $s_j$ pushes the model towards the prediction for $s_i$, whereas negative values imply competitive or adversarial influence.

FG-Trac logs these scores as part of the \textit{Sample Contribution Score Log}, binding them to operational checkpoints recorded by the Traceable Module. This transforms TracInCP from a retrospective estimator into a verifiable, temporally grounded mechanism for tracking how samples shape training dynamics.

\subsection{Blockchain Module}

Operational logs alone do not guarantee accountability. Without a protection mechanism, logs can be removed, edited, or regenerated after training, leaving no evidence that a modification occurred. In high-stakes machine learning systems, this is unacceptable: a trace is only trustworthy if it cannot be altered secretly. FG-Trac ensures this property through a lightweight anchoring mechanism that transforms operational logs into verifiable commitments.

FG-Trac batches logs into a Merkle tree, a hierarchical hashing structure. Intuitively, this structure works like a digital fingerprint. Each log entry contributes to the final fingerprint, and even a one-bit change anywhere in the logs produces a completely different fingerprint. Formally, given a log set $L=\{\ell_1,\dots,\ell_m\}$ and a collision-resistant hash function $H(\cdot)$, FG-Trac computes the Merkle root:
\begin{equation}
    r = H_{\mathrm{Merkle}}(L).
\end{equation}
The Merkle root $r$ acts as a concise, tamper-evident summary of all logs in $L$. Unlike storing logs directly, storing a single hash achieves two benefits:
\begin{itemize}
    \item Scalability: the size of $r$ itself is constant and does not depend on how many logs exist.
    \item Privacy preservation: raw logs remain off-chain; only their cryptographic fingerprint is public.
\end{itemize}

It is vital to distinguish between on-chain and off-chain costs. While the on-chain commitment stores only $r$ and therefore remains constant in size, the off-chain ledger that holds the log entries and intermediate Merkle nodes grows linearly with the number of recorded events. This separation allows FG-Trac to provide verifiable provenance without pushing large volumes of operational data to the blockchain.

FG-Trac periodically commits $r$ to a blockchain.
The blockchain functions as a public notary: once $r$ is recorded, no participant, including the developer, can overwrite it without leaving a trace, as each block stores both its own hash and the hash of the previous block. This creates an immutable sequence of commitments. Changing any log $\ell_k$ alters its hash, propagates through the Merkle tree, and produces a different root $\hat{r} \neq r$, immediately exposing tampering. Verification requires no trust in the pipeline operator: anyone holding a log entry and its Merkle proof can independently recompute $\hat{r}$ and verify consistency with the on-chain commitment.

In summary, the blockchain anchoring module converts FG-Trac’s operational logs from records that could be edited into evidence that cannot be denied. Without this component, FG-Trac would provide traceability, but not verifiable traceability.

\subsection{Query Module}

%FG-Trac exposes its operations to end users and auditors via a query interface. 
To address the third question: \emph{Can their influence be independently verified?} FG-Trac introduces a Query Module that enables users and auditors to validate sample-level logs and contribution scores.
While the Blockchain Module ensures that records are tamper-evident, the Query Module operationalises independent verification by retrieving authorised logs, providing their Merkle proofs, and recomputing a root hash to verify integrity.
Given a pseudonymous identifier $h_i$ and appropriate authorisation (e.g., a key or access token), the module performs three steps:
\begin{enumerate}
\item retrieves all logs associated with $h_i$ (user mapping, training role, training actions, modality attention, contribution scores, and prediction outcomes),
\item returns the corresponding Merkle proofs, and
\item recomputes a root $\hat{r}$ for verification.
\end{enumerate}

Verification succeeds if $\hat{r}$ matches the on-chain commitment $r$. Upon successful verification, FG-Trac automatically returns the authorised subset of logs associated with $h_i$, enabling users to inspect when their data was used, in what role, and how it influenced model behaviour. If verification fails, the system refuses to release the queried logs and alerts the user that the trace may have been tampered with. This enables users and auditors to answer operational questions such as: \textit{Was my data used? When, and in what role? How did it influence model behaviour? Has any log been altered since it was recorded?} The module is agnostic to the underlying model class and application domain. It applies to convolutional neural networks (CNNs), multimodal graph neural networks (GNNs), and other differentiable architectures deployed in high-risk settings.

The components introduced above operate jointly during pipeline execution. Algorithm~\ref{alg:fgtrac} summarises the FG-Trac workflow; detailed steps are deferred to the appendix for brevity. FG-Trac enables verifiable, sample-level traceability for machine learning pipelines whose ingestion, preprocessing, training, and inference stages are instrumentable and whose training process exposes gradients at selected checkpoints.

\begin{algorithm}[t]
\caption{FG-Trac End-to-End Procedure for Sample-Level Traceability}
\label{alg:fgtrac}
\begin{algorithmic}[1]
\Require Differentiable ML pipeline $\Pi$, dataset $\mathcal{D}$, checkpoint set $\mathcal{C}$ (gradients accessible at checkpoints)

\Statex \textbf{Phase 1: Sample Identification and Ingestion ($\mathcal{T}_0$)}
\For{sample $s_i \in \mathcal{D}$}
    \State $u_i \gets$ \Call{GetUserID}{$s_i$}
    \State $h_i \gets \mathrm{SHA256}(u_i)$ \Comment{Pseudonymous identifier}
    \State \Call{LogUserMapping}{$u_i, h_i$}
\EndFor

\Statex \textbf{Phase 2: Training-Time Instrumentation ($\mathcal{T}_1\!\rightarrow\!\mathcal{T}_2$)}
\For{epoch $e$ in $\Pi.\mathrm{training}$}
    \For{batch $B$ in $\Pi.\mathrm{loader}$}
        \State \Call{HookPreprocess}{$B$}
        \State \Call{HookTrainingActions}{$B$}
        \If{$\Pi$ \text{is multimodal}}
            \State \Call{HookModalityAttention}{$B$}
        \EndIf
    \EndFor

    \If{$e \in \mathcal{C}$}
        \State $c \gets e$
        \State $\theta_c \gets$ \Call{GetCheckpointParams}{$c$}
        \State \Call{ComputeContributionScores}{$\theta_c$}
        \State \Call{LogSampleContribution}{}
    \EndIf
\EndFor

\Statex \textbf{Phase 3: Tamper-Evident Anchoring ($\mathcal{T}_2\!\rightarrow\!\mathcal{T}_3$)}
\State $L \gets$ \Call{CollectRecentLogs}{}
\State $r \gets$ \Call{MerkleRoot}{$L$}
\State \Call{BlockchainCommit}{$r$}

\Statex \textbf{Phase 4: Audit Query and Verification}
\Function{Audit}{$h_i$}
    \State $L_i, P_i \gets$ \Call{RetrieveLogsAndProofs}{$h_i$}
    \State $r \gets$ \Call{GetOnChainCommitment}{}
    \State $\hat{r} \gets$ \Call{RecomputeRoot}{$L_i, P_i$}
    \If{$\hat{r} = r$}
        \State \Return $L_i$ \Comment{Verified trace}
    \Else
        \State \Return \textsc{TamperingDetected}
    \EndIf
\EndFunction

\end{algorithmic}
\end{algorithm}

\subsection{Computational Complexity}
We analyse the computational overhead introduced by FG-Trac relative to a standard training pipeline. Let $N$ denote the number of training samples, $|\mathcal{C}|$ the number of checkpoints, $|\theta|$ the parameter dimension, and $M$ the number of recorded log events.
The Traceable Module introduces $O(1)$ logging per sample, yielding $O(N)$ total overhead and $O(M)$ storage, with small constants since logs store only identifiers and summary statistics. The contribution estimator dominates the cost: computing a gradient for a single sample requires $O(|\theta|)$, and evaluating checkpointed influence scores for $K$ target samples scales as
\(
O(|\mathcal{C}| \cdot K \cdot N \cdot |\theta|),
\)
where $K \ll N$ in practice and $|\mathcal{C}|$ is typically small.
Provenance anchoring requires $O(B)$ hashing for a batch of $B$ logs, while on-chain storage remains $O(1)$ because only the Merkle root is committed; the off-chain ledger grows as $O(M)$.
Query verification is lightweight: log retrieval requires $O(\log M)$ lookups, and verifying $L_i$ logs costs $O(L_i \log M)$ via Merkle proofs. As $L_i$ is typically small, verification overhead is minimal. Overall, FG-Trac adds linear or logarithmic overhead to logging, anchoring, and verification, with influence estimation as the primary computational cost, preserving scalability while enabling verifiable sample-level traceability.

\section{Experiments}
\label{sec:experiments}

This section evaluates FG-Trac as an operational mechanism for verifiable, sample-level traceability in machine learning pipelines. The evaluation does not target accuracy gains, but examines whether FG-Trac preserves predictive performance while enabling verifiable, sample-level traceability.

To examine FG-Trac under heterogeneous workflow demands, we instantiate it in two contrasting machine learning pipelines: a conventional convolutional neural network (CNN) trained on a benchmark vision dataset for controlled instrumentation validation, and a multimodal graph-learning architecture applied to real-world clinical data. These complementary settings enable us to assess both the general applicability of FG-Trac’s instrumentation and its relevance in high-stakes settings where traceability and accountability are crucial.

\subsection{Experimental Setup}
\label{subsec:experimental-setup}

\paragraph{\textbf{Datasets.}}
We evaluate FG-Trac on three publicly available datasets spanning computer vision and clinical neuroimaging domains, summarised in Table~\ref{tab:dataset_summary}.

\begin{itemize}
    \item \textbf{CIFAR-10}~\cite{krizhevsky2009cifar10} consists of 60{,}000 natural RGB images of size 32$\times$32, evenly distributed across 10 object categories (e.g., airplane, cat, truck), with 6{,}000 images per class. The dataset is widely used as a canonical benchmark due to its balanced label distribution, compact image resolution, and absence of privacy-sensitive content.

    \item \textbf{ABIDE}~\cite{di2014autism} (Autism Brain Imaging Data Exchange) aggregates resting-state functional MRI (rs-fMRI) data from multiple international sites and includes 871 subjects. Each subject is represented by a 116$\times$116 functional connectivity (FC) matrix constructed using the Automated Anatomical Labeling (AAL) atlas, together with phenotypic attributes such as age, gender, IQ score, and imaging site. Its multi-site heterogeneity introduces nontrivial variability, making traceability and provenance tracking particularly relevant.

    \item \textbf{ADHD-200}~\cite{BELLEC2017275} contains rs-fMRI data for 582 subjects sourced from eight research sites. Each subject is encoded as a functional connectivity matrix derived from voxelwise signals, complemented by demographic variables including age, gender, and diagnostic metadata. Compared with ABIDE, ADHD-200 exhibits higher inter-site variability and motion-related artefacts, providing a challenging setting for evaluating trace logs under noisy neuroimaging conditions.
\end{itemize}

\begin{table}[t]
\centering
\caption{Summary statistics of datasets used in this study.}
\Description{Overview of datasets used in the experiments, including domain, learning task, number of samples with class distribution, and data modalities for CIFAR-10, ABIDE, and ADHD-200.}
\resizebox{\linewidth}{!}{%
\begin{tabular}{lccc}
\toprule
\textbf{Dataset} & \textbf{Domain} & \textbf{Samples} & \textbf{Modality} \\
\midrule
CIFAR-10  & Computer vision  & 60{,}000 & RGB images \\
ABIDE     & Clinical rs-fMRI & 871      & fMRI-derived graphs \\
ADHD-200  & Clinical rs-fMRI & 582      & fMRI-derived graphs \\
\bottomrule
\end{tabular}}
\label{tab:dataset_summary}
\end{table}

\paragraph{\textbf{Backbone models and baselines.}}
The evaluation considers two learning pipelines. For CIFAR-10, we adopt the standard CNN from the PyTorch tutorial~\cite{pytorch_cifar10_tutorial} as the backbone, evaluated with FG-Trac enabled and disabled. This comparison isolates the effect of sample-level traceability without introducing architectural changes.

For multimodal graph learning, FG-Trac is integrated into MM-GTUNets~\cite{10946209}, a representative model that jointly learns from functional connectivity matrices and phenotypic attributes. To situate this model within the broader landscape of neuroimaging graph learning, three established GNN methods are included as reference points: BrainGNN~\cite{LI2021102233}, GATE~\cite{9868044}, and MMGL~\cite{9733917}. These baselines are evaluated in their original form, while MM-GTUNets is examined both with and without FG-Trac. All models follow the same data preprocessing procedures, graph construction pipeline, and cross-validation protocol to ensure comparability.

\paragraph{\textbf{Implementation details.}}
All implementations are in PyTorch and run on Google Colaboratory (Runtime Version 2025.07) with Python 3.11.13 and an NVIDIA T4 GPU. The CIFAR-10 CNN is trained for 100 epochs with a batch size of 128 and a learning rate of $1\times10^{-3}$. MM-GTUNets is trained on ABIDE and ADHD-200 using 10-fold cross-validation with a fixed seed (911), a learning rate of $5\times10^{-4}$, the Adam optimiser, and early stopping after 20 non-improving epochs. After multimodal alignment, node features are 128-dimensional. Imaging and phenotypic subnetworks use three-layer U-Net encoders with 64 hidden channels and 128-dimensional outputs; dropout and edge-drop ratios are set to 0.3 and 0.2, respectively, and pooling ratios are $(0.5, 0.25)$. Graph regularisation and other optimisation hyperparameters follow~\cite{10946209}. All baseline models are trained using the identical configuration for fair comparison.

FG-Trac is evaluated using three complementary metric groups. \emph{Model-level} metrics (Accuracy, Sensitivity, Specificity, and AUC) verify that traceability does not affect predictive performance. \emph{System-level} metrics capture runtime and resource costs, including training time per epoch, inference latency, peak RAM and GPU usage, and blockchain ledger size. \emph{Trace-level} metrics assess output quality by reporting trace completeness, Merkle-root verification success, and TracInCP-derived contribution patterns, examining whether same-class samples exert supportive influence and cross-class samples exert opposing influence.

\subsection{Mechanism Validation on CIFAR-10}
\label{subsec:cifar10}

We first instantiate FG-Trac in a conventional vision setting using the PyTorch CIFAR-10 CNN as the backbone, which consists of three convolutional blocks followed by two fully connected layers. This experiment serves as a controlled, privacy-neutral testbed to validate the \emph{mechanics} of the framework: whether instrumentation leaves optimisation behaviour unchanged, whether end-to-end logging remains computationally feasible, and whether the recorded contribution scores reflect intuitive relationships among samples. CIFAR-10 is deliberately chosen not as a challenging benchmark, but as a simple, fully inspectable pipeline in which traceability can be exhaustively verified.

\begin{figure}[t]
    \centering
    \includegraphics[width=0.7\columnwidth]{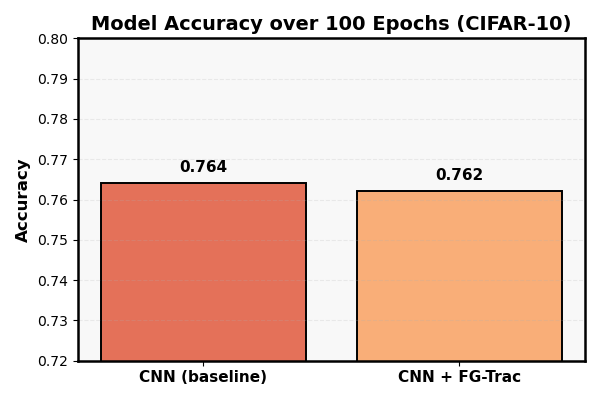}
    \caption{Classification accuracy comparison between baseline CNN and FG-Trac-integrated CNN on CIFAR-10 after 100 epochs.}
    \Description{Line plot comparing test classification accuracy over training epochs for a baseline CNN and the same CNN instrumented with FG-Trac on the CIFAR-10 dataset.}
    \label{fig:cnn_accuracy}
\end{figure}

As a sanity check, we compare the baseline CNN with its FG-Trac-instrumented counterpart after 100 epochs. The two models achieve almost identical test accuracies (0.764 vs.\ 0.762), as shown in (Figure~\ref{fig:cnn_accuracy}). Model performance is not the primary focus of this experiment; the goal is to demonstrate that enabling sample-level traceability does not disrupt the training dynamics of a standard CNN.

\begin{table}[t]
\centering
\caption{Runtime and memory usage of CNN models with and without FG-Trac.}
\label{tab:cnn-trace}
\resizebox{\linewidth}{!}{%
\begin{tabular}{lccc}
\toprule
\textbf{Model} & \textbf{Epoch Time (s)} & \textbf{RAM (MB)} & \textbf{Ledger (MB)} \\
\midrule
CNN (baseline)        & 9.31 $\pm$ 0.33 & 1655.1 & -- \\
CNN + \textbf{FG-Trac} & 9.47 $\pm$ 0.49 & 1712.2 & 1199 \\
\bottomrule
\end{tabular}}
\end{table}

\begin{figure}[t]
    \centering
    \includegraphics[width=\columnwidth]{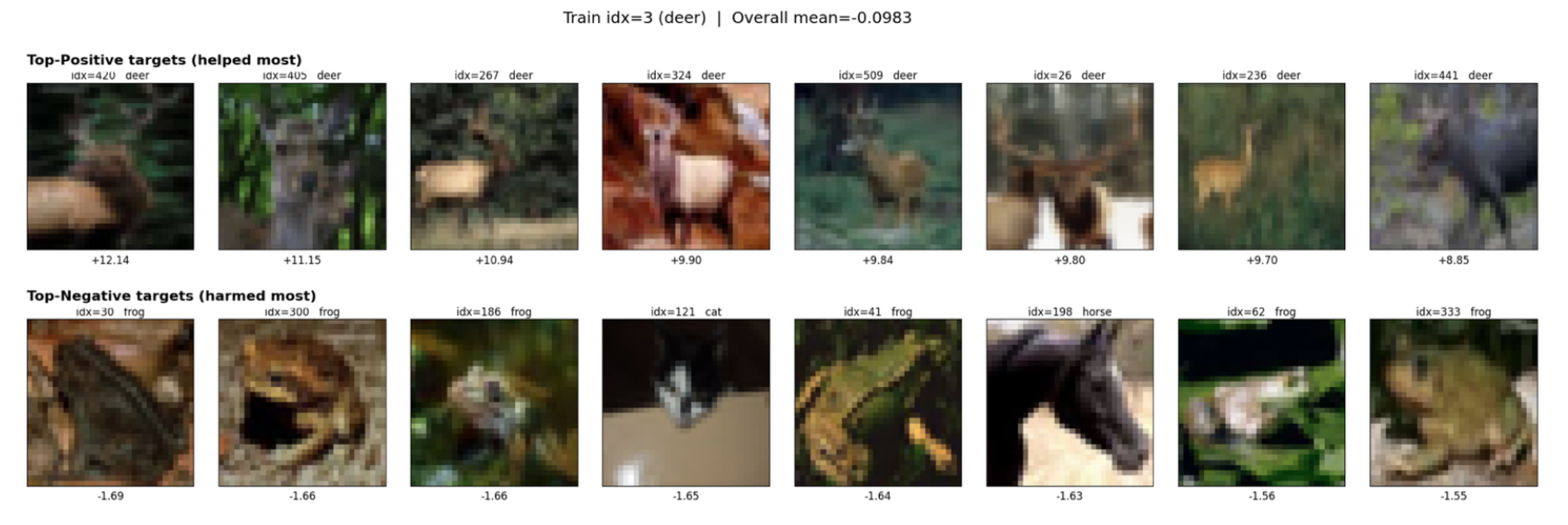}
    \caption{Visualisation of sample-level contribution reconstructed from a queried training sample.}
    \label{fig:cnn_case}
\end{figure}

From a systems perspective, the overhead of full logging is modest. As summarised in Table~\ref{tab:cnn-trace}, average epoch time increases from 9.31~s to 9.47~s (less than 2\%), and peak RAM rises from 1655.1~MB to 1712.2~MB. GPU memory consumption remains essentially unchanged because all trace hooks are implemented as asynchronous callbacks that write encrypted NDJSON records on the CPU while reusing the model's original forward and backward passes. After 100 epochs, the accumulated ledger reaches approximately 1.2~GB, reflecting complete off-chain logging of all sample-level events and the on-chain anchoring of Merkle roots. The overhead therefore scales linearly with the number of samples and training epochs and remains well within the capacity of commodity hardware.

%%%这个是gnn的图，排版问题才放在这里了。跟这部分内容没关系
\begin{table*}[t]
\centering
\caption{Classification performance on ABIDE and ADHD-200 datasets.}
\Description{Comparison of classification performance for multimodal graph learning models on ABIDE and ADHD-200, reported in terms of accuracy (ACC), sensitivity (SEN), specificity (SPE), and AUC.}
\resizebox{\textwidth}{!}{
\begin{tabular}{lcccccccc}
\toprule
\multirow{2}{*}{\textbf{Model}} &
\multicolumn{4}{c}{\textbf{ABIDE (HC vs ASD)}} &
\multicolumn{4}{c}{\textbf{ADHD-200 (HC vs ADHD)}} \\
\cmidrule(lr){2-5}\cmidrule(lr){6-9}
& ACC & SEN & SPE & AUC & ACC & SEN & SPE & AUC \\
\midrule
BrainGNN \cite{LI2021102233} 
& 71.30 $\pm$ 4.30 & 71.64 $\pm$ 3.97 & 70.26 $\pm$ 6.29 & 72.73 $\pm$ 4.29 
& 68.47 $\pm$ 5.53 & 65.36 $\pm$ 2.34 & 68.93 $\pm$ 3.67 & 73.41 $\pm$ 2.07 \\

GATE \cite{9868044}
& 76.56 $\pm$ 2.18 & 75.54 $\pm$ 3.85 & 72.63 $\pm$ 2.30 & 74.06 $\pm$ 3.69
& 71.01 $\pm$ 3.37 & 69.29 $\pm$ 2.17 & 74.85 $\pm$ 4.81 & 70.36 $\pm$ 3.16 \\

MMGL \cite{9733917}
& 76.98 $\pm$ 3.31 & 78.78 $\pm$ 5.19 & 74.61 $\pm$ 2.92 & 80.33 $\pm$ 2.33
& 74.41 $\pm$ 3.84 & 77.05 $\pm$ 6.62 & 76.14 $\pm$ 3.82 & 77.21 $\pm$ 4.69 \\
\midrule

MM-GTUNets \cite{10946209}
& 82.00 $\pm$ 0.20 & 84.40 $\pm$ 0.30 & \textbf{79.10 $\pm$ 1.10} & 88.70 $\pm$ 0.30
& 82.40 $\pm$ 0.60 & 89.60 $\pm$ 1.00 & \textbf{78.30 $\pm$ 0.90} & 91.00 $\pm$ 0.80 \\

MM-GTUNets + \textbf{FG-Trac}
& \textbf{82.00 $\pm$ 0.30} & \textbf{85.40 $\pm$ 0.30} & 78.00 $\pm$ 1.00 & \textbf{88.70 $\pm$ 0.40}
& \textbf{82.70 $\pm$ 0.60} & \textbf{91.80 $\pm$ 0.70} & 77.20 $\pm$ 0.80 & \textbf{91.20 $\pm$ 0.70} \\
\bottomrule
\end{tabular}}
\label{tab:performance}
\end{table*}

We then verify the correctness and integrity of the traceability. Using only the logs produced by FG-Trac, all 60{,}000 CIFAR-10 samples can be retrieved together with their T$_0$–T$_3$ event histories (ingestion, preprocessing, training, prediction). Recomputing Merkle roots from the NDJSON trace files yields hashes that exactly match the on-chain commitments for every run, indicating that no events have been lost or modified since anchoring. In effect, a conventional CNN training script becomes auditable at the granularity of individual data points once instrumented with FG-Trac.

Finally, we inspect the semantics of the recorded TracInCP contributions. By querying a randomly selected training example via its pseudonymous identifier and reconstructing its influence scores from the cached gradients, we obtain the contribution landscape shown in Figure~\ref{fig:cnn_case}. The exemplar, labelled as \emph{deer}, exhibits strongly positive influence on other \emph{deer} images (scores between +12.14 and +8.85) and weakly negative influence on samples from unrelated classes such as \emph{frog}, \emph{cat}, and \emph{horse} (scores around $-1.6$). This pattern is consistent with the intuition that within-class examples act as proponents, while cross-class examples act as mild opponents along the optimisation trajectory. Crucially, these relationships are reconstructed entirely from encrypted logs and gradient information: raw CIFAR-10 images are accessed post hoc only for visualisation, by matching pseudonymous identifiers against the public dataset. Together, these results confirm that FG-Trac can be plugged into a simple CNN pipeline to provide complete, verifiable and semantically meaningful sample-level traces with negligible impact on training.

This result supports the orthogonality of trace recording to the learning objective and confirms that FG-Trac operates transparently within the training workflow without altering predictive behaviour.

\subsection{Multimodal Graph Learning in a High-Risk Domain}
\label{subsec:exp-multimodal}

We next apply FG-Trac to a more demanding and clinically sensitive setting: multimodal graph-based prediction of mental health disorders on ABIDE and ADHD-200. In contrast to the CIFAR-10 experiment, this scenario involves heterogeneous neuroimaging and phenotypic modalities, graph construction over subject populations, attention-based fusion, and stricter privacy requirements. It therefore provides a realistic test of whether FG-Trac remains model-agnostic and operationally viable when deployed in high-risk machine learning pipelines.

\begin{table}[t]
\centering
\caption{Runtime of MM-GTUNets with and without FG-Trac (time in seconds).}
\Description{Comparison of training epoch time and inference time for MM-GTUNets with and without FG-Trac on the ABIDE and ADHD-200 datasets.}
\resizebox{\columnwidth}{!}{
\begin{tabular}{l|c|cc|cc}
\Xhline{1.2pt}
\multirow{2}{*}{Model} & \multirow{2}{*}{Trace} 
& \multicolumn{2}{c|}{ABIDE} 
& \multicolumn{2}{c}{ADHD-200} \\
\cline{3-6}
 & & Epoch & Inference & Epoch & Inference \\
\hline
MM-GTUNets                  & $\times$     & 0.5293 & 0.2353 & 0.2846 & 0.1132 \\
MM-GTUNets + \textbf{FG-Trac}        & $\checkmark$ & 1.3125 & 0.2366 & 0.8330 & 0.1132 \\
\Xhline{1.2pt}
\end{tabular}}
\label{runtime_comparison}
\end{table}

\begin{table}[t]
\centering
\caption{Resource usage of MM-GTUNets with and without FG-Trac (memory in MB).}
\resizebox{\columnwidth}{!}{
\begin{tabular}{l|c|ccc|ccc}
\Xhline{1.2pt}
\multirow{2}{*}{Model} & \multirow{2}{*}{Trace} 
& \multicolumn{3}{c|}{ABIDE} 
& \multicolumn{3}{c}{ADHD-200} \\
\cline{3-8}
 & & RAM & VRAM & Ledger & RAM & VRAM & Ledger \\
\hline
MM-GTUNets                  & $\times$     & 1587.39 & 4117.91 & --      & 1584.91 & 1874.80 & -- \\
MM-GTUNets + \textbf{FG-Trac}        & $\checkmark$ & 1715.54 & 4114.78 & 1072.17 & 1712.23 & 1875.10 & 478.76 \\
\Xhline{1.2pt}
\end{tabular}}
\label{tab:resource_comparison}
\end{table}

We integrate FG-Trac into MM-GTUNets, a multimodal graph transformer that jointly models functional connectivity matrices and demographic attributes~\cite{10946209}. Table~\ref{tab:performance} reports classification performance on ABIDE and ADHD-200 for three GNN baselines (BrainGNN, GATE, MMGL), MM-GTUNets without traceability, and MM-GTUNets with FG-Trac enabled. On both datasets, MM-GTUNets substantially outperform the baselines, and adding FG-Trac preserves this advantage: ACC and AUC remain essentially unchanged, and differences in SEN and SPE remain within 1 standard deviation. In some cases, the traceable model even exhibits slightly higher sensitivity (e.g., 91.8\% vs.\ 89.6\% on ADHD-200), while specificity varies only marginally. These results support the claim that FG-Trac can be overlaid on a state-of-the-art multimodal GNN without degrading diagnostic performance.

Runtime and resource statistics are summarised in Tables~\ref{runtime_comparison} and~\ref{tab:resource_comparison}. On ABIDE, average epoch time increases from 0.5293~s to 1.3125~s (a factor of 2.48), and on ADHD-200 from 0.2846~s to 0.8330~s (2.93$\times$), reflecting the cost of per-batch trace logging and checkpointed gradient caching for TracInCP on top of graph message passing and transformer layers. Importantly, inference latency is essentially unchanged (0.2353~s vs.\ 0.2366~s on ABIDE; 0.11316~s vs.\ 0.11320~s on ADHD-200), because prediction reuses the original forward path and does not emit logs or compute influence scores. Peak RAM usage increases by about 8\% due to encrypted NDJSON serialisation and Merkle-tree construction, whereas peak GPU memory remains almost identical, confirming that logging is handled off-GPU. The resulting blockchain-backed ledgers occupy around 1.1 GB for ABIDE and 0.48 GB for ADHD-200, scaling linearly with the number of subjects and training events. Overall, the cost of enabling full sample-level traceability in these multimodal graph pipelines is moderate and confined to the training phase, which can be scheduled offline in clinical workflows.

As in the CNN setting, we verify that FG-Trac provides complete and tamper-evident traces. For every subject in ABIDE and ADHD-200, the framework reconstructs the full lifecycle from data registration and split assignment (T$_0$–T$_1$), through multimodal graph construction and training interactions (T$_1$–T$_2$), to model-versioned predictions (T$_2$–T$_3$). Recomputed Merkle roots over all NDJSON logs match the on-chain commitments with no discrepancies, indicating that the trace ledger offers the same cryptographic integrity guarantees in a high-risk medical context as it does in the simple CIFAR-10 pipeline.

FG-Trac also enables contribution tracking in multimodal GNN pipelines without exposing raw clinical data. Using TracInCP, the framework computes per-sample influence statistics at selected checkpoints. It allows auditors to reconstruct these values on demand, confirming whether and how individual subjects participated in the optimisation process. In this high-privacy setting, we intentionally avoid interpreting contribution polarity with respect to diagnostic labels; FG-Trac provides mechanistic, auditable influence traces, rather than semantic explanations of clinical patterns.

\subsection{User-Level Auditing: End-to-End Traceability in Practice}
\label{subsec:case-study}

\begin{figure}[t]
    \centering
    \includegraphics[width=\columnwidth]{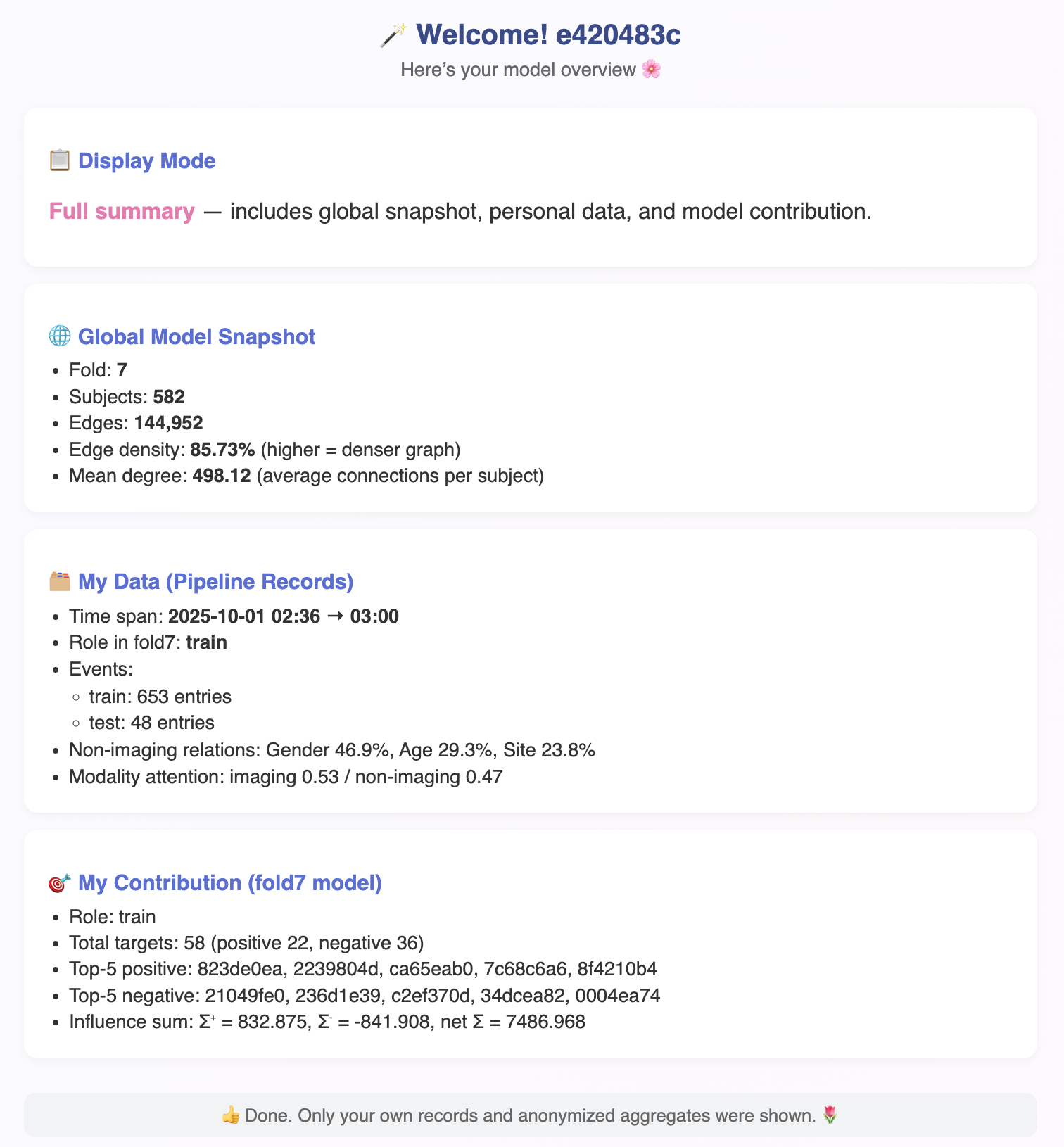}
    \caption{User query interface automatically generated from the FG-Trac auditing module for sample-level traceability}
    \label{fig:user_query}
\end{figure}

FG-Trac enables a new interaction modality: \emph{sample-level auditing}, where users can verify, through cryptographic evidence, whether and how their data contributed to a trained model. Figure~\ref{fig:user_query} shows the user-facing audit interface generated by FG-Trac, which exposes verified sample-level traces for authorised queries.

We illustrate this capability with a participant from ADHD-200, identified only by a pseudonymous hash $\text{hash}_{\text{id}}(u)$. Upon submitting this identifier to FG-Trac's Query and Verification Module, the system reconstructs three verifiable components of the participant’s audit trail:

\paragraph{(1) Participation Timeline.}
The framework discloses the subject’s lifecycle within the machine learning pipeline, including the assigned role (train/validation/test), timestamps of traceable events, and the specific training epoch in which the record was last used. Unlike aggregate provenance logs, the audit interface exposes concrete operational evidence, not merely that the dataset was used, but \emph{when and how this individual} was involved in the workflow.

\paragraph{(2) Modality Usage.}
For multimodal GNNs, FG-Trac reveals the attention weights allocated to imaging and phenotypic modalities when processing this subject's record. In the selected fold, the system reports an imaging-to-phenotype attention ratio of 0.53:0.47, indicating that both modalities contributed meaningfully to the final representation. This modality-level transparency would be inaccessible without trace hooks that capture operational signals during model execution.

\paragraph{(3) Contribution Summary.}
FG-Trac provides a signed influence profile derived from TracInCP. In this case, the subject exerts a positive influence on 22 peers and a negative impact on 36, yielding a net influence score (7486.968 in this run) aggregated over signed contributions. The top positively affected identifiers correspond to subjects within the same diagnostic group, reflecting clinically consistent behaviour. Importantly, each influence entry is accompanied by a Merkle proof, enabling independent verification against the on-chain commitment without exposing raw neuroimaging data.

This case study illustrates a capability absent from existing transparency mechanisms: individuals can trace their personal contribution to a model's behaviour and cryptographically verify that system logs have not been altered. FG-Trac provides an actionable auditing interface suitable for high-risk machine learning deployments where users, regulators, or organisations require verifiable guarantees of data usage.

Across both simple and high-risk pipelines, FG-Trac establishes verifiable, sample-level traceability with minimal perturbation to model behaviour. These results confirm that the framework is not tied to a particular model architecture, modality, or domain, and can serve as an operational transparency layer for modern machine learning systems.

\section{Conclusion}

This paper introduced FG-Trac, a model-agnostic framework for verifiable, sample-level traceability in machine learning pipelines. Rather than treating transparency as post-hoc documentation or interpretive analysis, FG-Trac instruments pipeline execution to record how individual data samples are processed and propagated across preprocessing, training, and prediction stages, anchoring these records to tamper-evident cryptographic commitments. By integrating checkpoint-resolved influence estimation with Merkle-based logging, FG-Trac enables independently verifiable audit trails without modifying model architectures or training objectives. Experiments on a conventional convolutional neural network and a multimodal graph neural network in clinical settings show that FG-Trac preserves predictive performance with practical and scalable overhead, demonstrating that fine-grained traceability can be embedded into modern machine learning workflows.

By transforming data-usage claims into verifiable operational evidence, FG-Trac provides a concrete auditing interface for high-risk machine learning deployments that require oversight, compliance, and post-hoc verification. FG-Trac assumes that pipeline execution can be instrumented as designed and does not address adversarial operators who intentionally bypass or suppress logging before commitment. Addressing such threats will require tighter integration with trusted execution environments, secure training infrastructures, or hardware-backed attestations. Future work will extend FG-Trac to distributed and federated pipelines and explore integration with policy-driven governance mechanisms, further strengthening operational accountability in complex Web and AI ecosystems.

%\begin{acks}
%To Robert, for the bagels and explaining CMYK and color spaces.
%\end{acks}

\bibliographystyle{ACM-Reference-Format}
\balance
\bibliography{web4good}

@String{Computing = "Computing" }

@article{tagliabue2021dag,
  title={DAG Card is the new Model Card},
  author={Tagliabue, Jacopo and Tuulos, Ville and Greco, Ciro and Dave, Valay},
  journal={arXiv preprint arXiv:2110.13601},
  year={2021}
}

@inproceedings{mitchell2019model,
  title={Model cards for model reporting},
  author={Mitchell, Margaret and Wu, Simone and Zaldivar, Andrew and Barnes, Parker and Vasserman, Lucy and Hutchinson, Ben and Spitzer, Elena and Raji, Inioluwa Deborah and Gebru, Timnit},
  booktitle={Proceedings of the conference on fairness, accountability, and transparency},
  pages={220--229},
  year={2019}
}

@article{chander2025toward,
  title={Toward trustworthy artificial intelligence (TAI) in the context of explainability and robustness},
  author={Chander, Bhanu and John, Chinju and Warrier, Lekha and Gopalakrishnan, Kumaravelan},
  journal={ACM Computing Surveys},
  volume={57},
  number={6},
  pages={1--49},
  year={2025},
}

@article{pruthi2020estimating,
  title={Estimating training data influence by tracing gradient descent},
  author={Pruthi, Garima and Liu, Frederick and Kale, Satyen and Sundararajan, Mukund},
  journal={Advances in Neural Information Processing Systems},
  volume={33},
  pages={19920--19930},
  year={2020}
}

@misc{pytorch_cifar10_tutorial,
  title        = {Training a classifier},
  author       = {{PyTorch Team}},
  year         = {2024},
  howpublished = {\url{https://pytorch.org/tutorials/beginner/blitz/cifar10_tutorial.html}},
  note         = {Accessed: 2025-10-09}
}

@ARTICLE{10946209,
  author={Cai, Luhui and Zeng, Weiming and Chen, Hongyu and Zhang, Hua and Li, Yueyang and Feng, Yu and Yan, Hongjie and Bian, Lingbin and Ting Siok, Wai and Wang, Nizhuan},
  journal={IEEE Transactions on Medical Imaging}, 
  title={MM-GTUNets: Unified Multi-Modal Graph Deep Learning for Brain Disorders Prediction}, 
  year={2025},
  volume={44},
  number={9},
  pages={3705-3716},}

@inproceedings{koh2017understanding,
  title={Understanding black-box predictions via influence functions},
  author={Koh, Pang Wei and Liang, Percy},
  booktitle={International Conference on Machine Learning},
  pages={1885--1894},
  year={2017},
  organization={PMLR}
}

@inproceedings{jia2019towards,
  title={Towards efficient data valuation based on the shapley value},
  author={Jia, Ruoxi and Dao, David and Wang, Boxin and Hubis, Frances Ann and Hynes, Nick and G{\"u}rel, Nezihe Merve and Li, Bo and Zhang, Ce and Song, Dawn and Spanos, Costas J},
  booktitle={The 22nd International Conference on Artificial Intelligence and Statistics},
  pages={1167--1176},
  year={2019},
  organization={PMLR}
}

@inproceedings{krizhevsky2009cifar10,
 author = {Srivastava, Rupesh K and Greff, Klaus and Schmidhuber, J\"{u}rgen},
 booktitle = {Advances in Neural Information Processing Systems},
 title = {Training Very Deep Networks},
 volume = {28},
 year = {2015}
}

@article{di2014autism,
  title={The autism brain imaging data exchange: towards a large-scale evaluation of the intrinsic brain architecture in autism},
  author={Di Martino, Adriana and Yan, Chao-Gan and Li, Qingyang and Denio, Erin and Castellanos, Francisco X and Alaerts, Kaat and Anderson, Jeffrey S and Assaf, Michal and Bookheimer, Susan Y and Dapretto, Mirella and others},
  journal={Molecular psychiatry},
  volume={19},
  number={6},
  pages={659--667},
  year={2014},
}

@article{BELLEC2017275,
title = {The Neuro Bureau ADHD-200 Preprocessed repository},
journal = {NeuroImage},
volume = {144},
pages = {275-286},
year = {2017},
author = {Pierre Bellec and Carlton Chu and François Chouinard-Decorte and Yassine Benhajali and Daniel S. Margulies and R. Cameron Craddock},
}

@article{LI2021102233,
title = {BrainGNN: Interpretable Brain Graph Neural Network for fMRI Analysis},
journal = {Medical Image Analysis},
volume = {74},
pages = {102233},
year = {2021},
author = {Xiaoxiao Li and Yuan Zhou and Nicha Dvornek and Muhan Zhang and Siyuan Gao and Juntang Zhuang and Dustin Scheinost and Lawrence H. Staib and Pamela Ventola and James S. Duncan},
}

@ARTICLE{9868044,
  author={Peng, Liang and Wang, Nan and Xu, Jie and Zhu, Xiaofeng and Li, Xiaoxiao},
  journal={IEEE Transactions on Medical Imaging}, 
  title={GATE: Graph CCA for Temporal Self-Supervised Learning for Label-Efficient fMRI Analysis}, 
  year={2023},
  volume={42},
  number={2},
  pages={391-402},
}

@article{9733917,
	author = {Zheng, Shuai and Zhu, Zhenfeng and Liu, Zhizhe and Guo, Zhenyu and Liu, Yang and Yang, Yuchen and Zhao, Yao},
	journal = {IEEE Transactions on Medical Imaging},
	number = {9},
	pages = {2207-2216},
	title = {Multi-Modal Graph Learning for Disease Prediction},
	volume = {41},
	year = {2022},}

@article{mora2021traceability,
  title={Traceability for trustworthy AI: a review of models and tools},
  author={Mora-Cantallops, Mar{\c{c}}al and S{\'a}nchez-Alonso, Salvador and Garc{\'\i}a-Barriocanal, Elena and Sicilia, Miguel-Angel},
  journal={Big Data and Cognitive Computing},
  volume={5},
  number={2},
  pages={20},
  year={2021},
  publisher={MDPI}
}

@article{huang2025instance,
  title={Instance-Level Data-Use Auditing of Visual ML Models},
  author={Huang, Zonghao and Gong, Neil Zhenqiang and Reiter, Michael K},
  journal={ArXiv Preprint ArXiv:2503.22413},
  year={2025}
}

@article{schlegel2025capturing,
  title={Capturing end-to-end provenance for machine learning pipelines},
  author={Schlegel, Marius and Sattler, Kai-Uwe},
  journal={Information Systems},
  volume={132},
  pages={102495},
  year={2025},
  publisher={Elsevier}
}

@article{spoczynski2025atlas,
  title={Atlas: A framework for ml lifecycle provenance \& transparency},
  author={Spoczynski, Marcin and Melara, Marcela S and Szyller, Sebastian},
  journal={ArXiv Preprint ArXiv:2502.19567},
  year={2025}
}

@inproceedings{grafberger2023provenance,
  title={Provenance tracking for end-to-end machine learning pipelines},
  author={Grafberger, Stefan and Groth, Paul and Schelter, Sebastian},
  booktitle={Companion Proceedings of the ACM Web Conference 2023},
  pages={1512--1512},
  year={2023}
}

@article{mu2023data,
  title={Data provenance via differential auditing},
  author={Mu, Xin and Pang, Ming and Zhu, Feida},
  journal={IEEE Transactions on Knowledge and Data Engineering},
  volume={36},
  number={10},
  pages={5066--5079},
  year={2023},
  publisher={IEEE}
}

@article{hupont2024use,
  title={Use case cards: a use case reporting framework inspired by the European AI Act},
  author={Hupont, Isabelle and Fern{\'a}ndez-Llorca, David and Baldassarri, Sandra and G{\'o}mez, Emilia},
  journal={Ethics and Information Technology},
  volume={26},
  number={2},
  pages={19},
  year={2024},
}

@inproceedings{pushkarna2022data,
  title={Data cards: Purposeful and transparent dataset documentation for responsible ai},
  author={Pushkarna, Mahima and Zaldivar, Andrew and Kjartansson, Oddur},
  booktitle={Proceedings of the 2022 ACM Conference on Fairness, Accountability, and Transparency},
  pages={1776--1826},
  year={2022}
}

@article{sui2021representer,
  title={Representer point selection via local jacobian expansion for post-hoc classifier explanation of deep neural networks and ensemble models},
  author={Sui, Yi and Wu, Ga and Sanner, Scott},
  journal={Advances in neural information processing systems},
  volume={34},
  pages={23347--23358},
  year={2021}
}

@inproceedings{souza2025prov,
  title={PROV-AGENT: Unified provenance for tracking AI agent interactions in agentic workflows},
  author={Souza, Renan and Gueroudji, Amal and DeWitt, Stephen and Rosendo, Daniel and Ghosal, Tirthankar and Ross, Robert and Balaprakash, Prasanna and Da Silva, Rafael Ferreira},
  booktitle={2025 IEEE International Conference on eScience (eScience)},
  pages={467--473},
  year={2025},
}

@inproceedings{chen2020developments,
  title={Developments in mlflow: A system to accelerate the machine learning lifecycle},
  author={Chen, Andrew and Chow, Andy and Davidson, Aaron and DCunha, Arjun and Ghodsi, Ali and Hong, Sue Ann and Konwinski, Andy and Mewald, Clemens and Murching, Siddharth and Nykodym, Tomas and others},
  booktitle={Proceedings of the fourth international workshop on data management for end-to-end machine learning},
  pages={1--4},
  year={2020}
}

@inproceedings{vartak2016modeldb,
  title={ModelDB: a system for machine learning model management},
  author={Vartak, Manasi and Subramanyam, Harihar and Lee, Wei-En and Viswanathan, Srinidhi and Husnoo, Saadiyah and Madden, Samuel and Zaharia, Matei},
  booktitle={Proceedings of the Workshop on Human-In-the-Loop Data Analytics},
  pages={1--3},
  year={2016}
}

@article{casalicchio2019openml,
  title={OpenML: An R package to connect to the machine learning platform OpenML},
  author={Casalicchio, Giuseppe and Bossek, Jakob and Lang, Michel and Kirchhoff, Dominik and Kerschke, Pascal and Hofner, Benjamin and Seibold, Heidi and Vanschoren, Joaquin and Bischl, Bernd},
  journal={Computational Statistics},
  volume={34},
  number={3},
  pages={977--991},
  year={2019},
}

@inproceedings{10.1145/3583780.3615998,
	author = {Boukharouba, Ikram and S\`{e}des, Florence and Bortolaso, Christophe and Mouysset, Florent},
	booktitle = {Proceedings of the 32nd ACM International Conference on Information and Knowledge Management},
	doi = {10.1145/3583780.3615998},
	numpages = {2},
	pages = {5236--5237},
	series = {CIKM '23},
	title = {From User Activity Traces to Navigation Graph for Software Enhancement: An Application of Graph Neural Network (GNN) on a Real-World Non-Attributed Graph},
	year = {2023},}

@article{liu2025trace2vec,
	author = {Liu, Wei and Gao, Peng and Zhang, Haotian and Li, Ke and Yang, Weiyong and Wei, Xingshen and Shu, Jiwu},
	journal = {Pattern Recognition},
	pages = {111363},
	title = {Trace2Vec: Detecting complex multi-step attacks with explainable graph neural network},
	year = {2025}}

@article{yanez2025text,
	author = {Y{\'a}{\~n}ez-Romero, Fabio and Montoyo, Andr{\'e}s and Su{\'a}rez, Armando and Guti{\'e}rrez, Yoan and Mitkov, Ruslan},
	journal = {arXiv preprint arXiv:2504.02064},
	title = {From Text to Graph: Leveraging Graph Neural Networks for Enhanced Explainability in NLP},
	year = {2025}}

@article{li2023graph,
	author = {Li, Zhiyuan and He, Enhan},
	journal = {IEEE Access},
	pages = {62109--62120},
	publisher = {IEEE},
	title = {Graph neural network-based bitcoin transaction tracking model},
	volume = {11},
	year = {2023}}

@article{10185966,
	author = {Cai, Jiahong and Liang, Wei and Li, Xiong and Li, Kuanching and Gui, Zhenwen and Khan, Muhammad Khurram},
	doi = {10.1109/JIOT.2023.3296469},
	journal = {IEEE Internet of Things Journal},
	number = {24},
	pages = {21502-21514},
	title = {GTxChain: A Secure IoT Smart Blockchain Architecture Based on Graph Neural Network},
	volume = {10},
	year = {2023},}


\appendix
\section{Checkpoint Selection and Gradient Extraction}
\label{appendix:checkpoint}

FG-Trac computes sample-level contribution scores using a checkpointed variant of TracInCP, which requires access to gradients at selected optimisation states. This appendix describes the checkpoint selection strategy and the minimal gradient extraction used in our experiments.

\paragraph{Checkpoint Selection.}
Rather than storing gradients at every training step, FG-Trac monitors validation loss and selects a small set of checkpoints $\mathcal{C}$ corresponding to substantial decreases beyond random performance. These checkpoints capture meaningful optimisation progress while avoiding early noisy gradients and late flat loss regions. In all experiments, we use $|\mathcal{C}|=3$, which provides a practical trade-off between computational overhead and interpretability.

\paragraph{Gradient Extraction.}
At each checkpoint $c \in \mathcal{C}$, FG-Trac caches the gradient of the task loss with respect to the final prediction layer parameters:
\[
\nabla_{\theta}\ell(\theta_c, s) \quad \forall s \in \text{train set}.
\]
Restricting extraction to the final layer (as in TracInCP~\cite{pruthi2020estimating}) reduces storage and runtime costs while retaining a practical signal of how samples support or oppose downstream predictions. These cached gradients are reused to compute signed influence scores:
\[
I_{\mathrm{CP}}(s_j \!\rightarrow\! s_i)
=
\sum_{c\in\mathcal{C}}
\nabla_{\theta}\ell(\theta_c, s_i)^{\!\top}
\nabla_{\theta}\ell(\theta_c, s_j).
\]
This enables reconstruction of influence traces aligned with the optimisation trajectory without replaying training or accessing intermediate activations.

\end{document}